\documentclass[sigconf]{acmart}
\usepackage{balance}
\usepackage{natbib}
\usepackage{mathtools}
\usepackage{multicol, multirow}
\usepackage{threeparttable}
\usepackage{booktabs}
\usepackage{amsmath}

\DeclareMathOperator*{\argmin}{arg\,min}

\usepackage{subfigure}
\usepackage{algorithm}
\usepackage[noend]{algpseudocode}
\usepackage{ulem}
\usepackage{caption}
\usepackage{bbding}
\usepackage{enumitem}
\usepackage{dutchcal}

\makeatletter
\def\algbackskip{\hskip-\ALG@thistlm}
\makeatother

\renewcommand{\argmin}{\text{arg}\min}


\copyrightyear{2025}
\acmYear{2025}
\setcopyright{acmlicensed}\acmConference[KDD '25]{Proceedings of the 31st ACM SIGKDD Conference on Knowledge Discovery and Data Mining V.2}{August 3--7, 2025}{Toronto, ON, Canada}
\acmBooktitle{Proceedings of the 31st ACM SIGKDD Conference on Knowledge Discovery and Data Mining V.2 (KDD '25), August 3--7, 2025, Toronto, ON, Canada}
\acmDOI{10.1145/3711896.3737268}
\acmISBN{979-8-4007-1454-2/2025/08}

\begin{document}

\title{Timing is Important: Risk-aware Fund Allocation based on Time-Series Forecasting}

\author{Fuyuan Lyu}
\affiliation{
  \institution{McGill University \&\\ MILA - Quebec AI Institute}
  \city{Montreal}
  \country{Canada}
}
\email{fuyuan.lyu@mail.mcgill.ca}
\orcid{0000-0001-9345-1828}

\author{Linfeng Du}
\affiliation{
  \institution{McGill University}
  \city{Montreal}
  \country{Canada}
}
\email{linfeng.du@mail.mcgill.ca}
\orcid{0000-0002-4389-5575}

\author{Yunpeng Weng}
\affiliation{
  \institution{FiT, Tencent}
  \city{Shenzhen}
  \country{China}
}
\email{edwinweng@tencent.com}
\orcid{0000-0001-7593-2169}

\author{Qiufang Ying}
\affiliation{
  \institution{FiT, Tencent}
  \city{Shenzhen}
  \country{China}
}
\orcid{0000-0002-5298-2603}

\author{Zhiyan Xu}
\affiliation{
  \institution{FiT, Tencent}
  \city{Shenzhen}
  \country{China}
}
\orcid{0009-0004-7764-7803}

\author{Wen Zou}
\affiliation{
  \institution{FiT, Tencent}
  \city{Shenzhen}
  \country{China}
}
\orcid{0009-0001-1319-2203}

\author{Haolun Wu}
\affiliation{
  \institution{McGill University \&\\ MILA - Quebec AI Institute}
  \city{Montreal}
  \country{Canada}
}
\email{haolun.wu@mail.mcgill.ca}
\orcid{0000-0001-6255-1535}

\author{Xiuqiang He}
\affiliation{
  \institution{College of Big Data and Internet\\Shenzhen Technology University}
  \city{Shenzhen}
  \country{China}
}
\authornote{Corresponding Author}
\email{he.xiuqiang@gmail.com}
\orcid{0000-0002-4115-8205}

\author{Xing Tang}
\affiliation{
  \institution{College of Big Data and Internet\\Shenzhen Technology University}
  \city{Shenzhen}
  \country{China}
}
\email{xing.tang@hotmail.com}
\orcid{0000-0003-4360-0754}

\renewcommand{\shortauthors}{Fuyuan Lyu et al.}

\begin{abstract}
Fund allocation has been an increasingly important problem in the financial domain.
In reality, we aim to allocate the funds to buy certain assets within a certain future period. 
Naive solutions such as prediction-only or Predict-then-Optimize approaches suffer from goal mismatch.
Additionally, the introduction of the SOTA time series forecasting model inevitably introduces additional uncertainty in the predicted result.
To solve both problems mentioned above, we introduce a \textbf{R}isk-aware \textbf{T}ime-\textbf{S}eries \textbf{P}redict-a\textbf{n}d-\textbf{A}llocate (RTS-PnO) framework, which holds no prior assumption on the forecasting models.
Such a framework contains three features: (i) end-to-end training with objective alignment measurement, (ii) adaptive forecasting uncertainty calibration, and (iii) agnostic towards forecasting models. 
The evaluation of RTS-PnO is conducted over both online and offline experiments. For offline experiments, eight datasets from three categories of financial applications are used: Currency, Stock, and Cryptos. RTS-PnO consistently outperforms other competitive baselines. The online experiment is conducted on the Cross-Border Payment business at FiT, Tencent, and an 8.4\% decrease in regret is witnessed when compared with the product-line approach.
The code for the offline experiment is available here\footnote{https://github.com/fuyuanlyu/RTS-PnO}.
\end{abstract}




\maketitle
\section{Introduction}
\label{sec:introduction}

Fund Allocation has been an increasingly important problem in financial technology. Proper fund allocation can reduce the cost of financial operations. Previous studies allocate funds across different assets, such as stocks and currencies~\cite{FA1}. However, one commonly overlooked aspect is that the price of assets tends to vary rapidly over time, making the timing of acquiring assets equally important. This paper aims to investigate the fund allocation throughout the time dimension. Specifically, the goal is to acquire a certain amount of assets at minimal cost over a period of time.


One naive solution for asset allocation is to adopt SOTA time-series (TS) forecasting models~\cite{DLinear, PatchTST} directly to predict the price of the target asset over a period of time and heuristically select the lowest point, given that the price of assets is sequentially ordered as a time series. However, one fundamental problem of such a solution lies in the external constraints. Financial markets usually have auxiliary regulations such as risk management or internal controls. Hence, the lowest point is not necessarily feasible in certain cases. 

To deliver a feasible and accountable action, 
a Predict-then-Optimize (PtO) framework is commonly introduced~\cite{PTOCA,SPOTs,PtO4MDP}. The PtO framework intuitively decomposes the task into two sequential steps: predict and optimize. After obtaining the forecasting result under the supervision of future values, it subsequently utilizes off-the-shelf commercial solvers, such as Gurobi~\cite{Gurobi} or COPT~\cite{copt}, to obtain solutions with given constraints. Such a design is intuitively based on the hypothesis that a higher prediction accuracy (measured by prediction metrics such as \textit{MSE}) would result in better decision quality (measured by decision metrics such as \textit{Regret}) with theoretical support~\cite{PtO-bound}. The effectiveness of the PtO paradigm has been successfully demonstrated in previous industrial applications, such as courier allocation~\cite{PTOCA}.

However, adopting the PtO paradigm has certain challenges that have not been resolved. 
First, both empirical~\cite{PTO-PNO-Benchmark} and theoretical~\cite{SPO+,PnO-bound} gaps are witnessed between prediction objectives and business decision goals. The time-series forecasting models aim to predict future prices across all time stamps accurately. In other words, the value of all time stamps contributes equally to the training loss, e.g., MSE. In contrast, the optimization process tends to care more about extreme cases, such as the minimal or maximal values. By solving them in two subsequent steps, the PtO paradigm could eventually lead to suboptimal decisions. 
Additionally, the frequencies of financial data tend to be higher (minute-level), eventually making them harder to predict~\cite{neurips24talk}.
Second, such a paradigm overlooks the uncertainty of forecasting. The consequences of a bad financial decision could lead to huge losses for both the company and the customer. 
Hence, proposing an uncertainty measurement approach that correlates directly with the forecasted result is important. However, these methods tend to suffer from 
Previous research tends to rely on probabilistic models forecasting both the result and its uncertainty~\cite{DeepAR,D3VAE}. However, these methods suffer from cumulative errors as they decode each time step recursively. This drawback makes them unsuitable for real-world tasks requiring a long forecasting period.
Therefore, an uncertainty quantification measure that is model-agnostic and can forecast long periods directly is desired. 


To solve the two problems mentioned above in a unified approach, we propose a risk-aware time-series forecasting Predict-and-Allocate (RTS-PnO) framework to solve the fund allocation problem in the time domain. The RTS-PnO framework features three things: (i) end-to-end trainable with objective alignment, (ii) adaptive uncertainty measurement, and (iii) agnostic towards forecasting models. 
To alleviate the objective mismatch between prediction and optimization, RTS-PnO adopts the recently proposed Predict-and-Optimize (PnO) paradigm~\cite{SPO+}, also known as decision-focused learning (DFL). The PnO paradigm directly trains the forecasting model with surrogate losses approximating the feedback from the optimization stage. Though a decrease in prediction accuracy is witnessed under certain cases, an increase in the decision quality can be witnessed~\cite{PTO-PNO-Benchmark}. 
Additionally, inspired by the success of conformal prediction and its extension in the time series domain~\cite{CF-RNN,EnbPI,EnbPI2}, we propose to measure the forecasting uncertainty adaptively during the training loop. Such a design can iteratively calibrate the uncertainty condition in the surrogate problem, yielding a better decision.
Finally, all the designs of RTS-PnO do not make any prior assumptions about the architecture of forecasting models. Therefore, it is easy to update the forecasting models with advanced ones in the future.

Our method is evaluated in both offline and online experiments. For offline experiments, we select eight public benchmarks from three categories of financial applications. RTS-PnO consistently yields better performances than other baselines. We deploy RTS-PnO on Tencent’s financial platform for the online experiment to support cross-border payment scenarios for WeChat Pay. An average regret reduction of 8.4\% is witnessed over 12 time slots compared with the product-line baseline. 
To sum up, our contributions can be summarized as follows:
\begin{itemize}[topsep=0pt,noitemsep,nolistsep,leftmargin=*]
    \item We first study the fund allocation over the time domain, where the prices of assets vary over the time dimension.
    \item To align the training objective of time-series forecasting models and business criteria, we propose two model-agnostic frameworks, named RTS-PtO and RTS-PnO. RTS-PnO adopts an end-to-end training paradigm driven by the final objective and adaptively calibrates the uncertainty constraint during the process. RTS-PtO adopts a two-stage solution by training a prediction model and then solving the optimization with the fixed uncertainty constraint.
    \item Extensive evaluation is conducted on both offline benchmarks and online scenarios. Both results prove the effectiveness of the proposed frameworks.
\end{itemize}


\section{Related Work}
\label{sec:rw}

\subsection{Time Series Forecasting}

Modern architectures for time series forecasting aim to extend the forecasting horizon and improve long-term accuracy. Inspired by the success of Transformer-based models in capturing long-range dependencies, researchers have explored various adaptations of the Transformer architecture for this task. These include i) reducing computational complexity to sub-quadratic levels using sparse~\cite{Informer} and hierarchical~\cite{Pyraformer} attention, ii) extending the attention mechanism’s point-wise dependency modeling to capture segment-wise~\cite{LogSparse} and patch-wise dependencies~\cite{PatchTST, Crossformer}, and iii) modifying the attention mechanism to incorporate domain-specific processing techniques~\cite{Autoformer, FEDformer}. Besides Transformer-based models, modern temporal convolutional networks have also been shown to achieve competitive performance. MICN~\cite{MICN} combines local and global convolutions to better model long sequences, while TimesNet~\cite{TimesNet} reshapes the 1D series into 2D matrices based on salient periodicities to jointly model intra-period and inter-period variations. In fact, with the recent rise of linear models~\cite{DLinear} and MLPs~\cite{TSMixer}, the de facto neural architecture for this task remains undecided. In this work, we demonstrate the wide compatibility of RTS-PtO and RTS-PnO across various model architectures.

One drawback of the above-mentioned methods is the lack of uncertainty quantification. Existing approaches resort to generative modeling~\cite{DeepAR, D3VAE}, which naturally captures data variation. However, these approaches are often limited to short-term prediction, as modeling the joint data probability becomes exponentially difficult. Alternatively, we leverage the conformal prediction framework to characterize uncertainty for longer series~\cite{CF-RNN,EnbPI,EnbPI2}, which we show empirically can help achieve satisfactory performance across different datasets.

\subsection{From PtO To PnO}

The predict-then-optimize (PtO) can be viewed as an abstractive problem for many real-world applications, such as portfolio management or power scheduling, requiring both predicting unknown values and optimizing the target given these unknown values~\cite{Prescriptive,PtO-bound}. Such a paradigm has been recently extended to other large-scale applications, such as carrier allocation~\cite{PTOCA}, fund recommendation~\cite{PTOFA}.
However, it is believed that a misalignment of targets exists between prediction and optimization stages. Researchers are increasingly interested in training the prediction model directly targeting the optimization goal, commonly known as predict-and-optimize (PnO)~\cite{PTO-PNO-Benchmark,PtOorPnO,PnO-bound} or decision-focused learning~\cite{DFL-Survey}. The core challenge is to obtain meaningful gradients for model updating, given the optimization stage. Certain researchers adopt analytical approaches and aim to make the optimization layer differentible~\cite{OptNet,Cvxpylayers}. However, these works tend to rely on strong requirements on the objective functions or constraints, restricting their application scopes in reality. Other researchers~\cite{NCE,SPO+} instead adopt surrogate loss for the optimization layer and prove its convergence both theoretically and empirically. Our RTS-PnO first extends the application of the predict-and-optimization paradigm to large-scale industrial problems.

\section{Methodology}
\label{sec:method}

In this section, we introduce the problem formulation of the fund allocation over time in Section~\ref{sec:method_formulate}. Section~\ref{sec:method_pto} details the two-stage solution RTS-PtO based on predict-then-optimize. Section~\ref{sec:method_rtspno} details the end-to-end approach RTS-PnO.

\subsection{Problem Formulation}
\label{sec:method_formulate}

In this section, we propose the formal formulation of fund allocation over the time dimension. Without loss of generality, we focus on univariate series. The formulation for multi-variate time series can be derived easily. Suppose the unit price of a certain asset at time step $t$ is defined as $p_t$, then the unit price of that asset can formulate a time series, denoted as:
$$
\underbrace{p_1, p_2, \cdots, p_{t-1}, p_t}_{\text{known}} \ | \ \underbrace{p_{t+1}, p_{t+2}, \cdots}_{\text{unknown}}
$$
The goal of fund allocation over time is to acquire a certain amount of asset in $H$ future time steps $\left[p_{t+1}, p_{t+2}, \cdots, p_{t+H}\right]$, at the lowest cost. This can be formulated as:
$$\min \ \mathbf{a} \times \left[p_{t+1}, p_{t+2}, \cdots, p_{t+H}\right].$$
Here $\mathbf{a} \in \mathcal{A}$ denotes the allocation results, and $\mathcal{A} \subseteq [0, 1]^{H}$ represents the feasible allocation space. Again, we can assume the unit amplitude assumption on $\mathbf{a}$, denoted as $\sum \mathbf{a} = 1$. Therefore, the final objective can be formulated as:
\begin{equation}
\label{eq:formulation}
\begin{aligned}
    & \min \ \mathbf{a} \times \left[p_{t+1}, p_{t+2}, \cdots, p_{t+H}\right]. \\
    & s.t. \ \sum \mathbf{a} = 1, \mathbf{a} \in \mathcal{A}, \mathcal{A} \subseteq [0, 1]^{H}.
\end{aligned}
\end{equation}

Although the future price $\left[p_{t+1}, p_{t+2}, \cdots, p_{t+H}\right]$ has ground truth values, these values are unknown by the time $t$ when we make the allocation. Therefore, we need to forecast the future price and denote the predicted result as $\left[\hat{p}_{t+1}, \hat{p}_{t+2}, \cdots, \hat{p}_{t+H}\right]$. Hence, we propose two solutions to solve the above problem. First is a Predict-then-Optimize (PtO) framework, which treats both $\mathbf{a}$ and $\left[\hat{p}_{t+1}, \hat{p}_{t+2}, \cdots, \hat{p}_{t+H}\right]$ as independent variables. It makes the prediction first, then optimizes $\mathbf{a}$ given the predicted result. Second is a Predict-and-Optimize (PnO) framework, which treats $\mathbf{a}$ as a function of $\left[\hat{p}_{t+1}, \hat{p}_{t+2}, \cdots, \hat{p}_{t+H}\right]$ and conducts prediction and optimization simultaneously.

\subsection{RTS-PtO: A Two-Stage Solution with Uncertainty Constraints}
\label{sec:method_pto}

In this section, we first introduce the Risk-aware Predict-then-Optimize (PtO) framework, which is commonly adopted by similar problems~\cite{PTOCA}. The PtO solution naturally consists of two steps: (i) predicting the future price of the asset given the historical records and contextual information, and (ii) obtaining the allocation result based on the predicted price. The first step, like other forecasting problems, aims at accurately predicting the price in the future. The second step, on the other hand, can be viewed as solving an optimization problem targeting minimal cost reduction under constraints. Additionally, we propose an additional uncertainty constraint on the forecasted results to avoid over-aggressive decisions.

\subsubsection{Prediction Stage}
During the first forecasting stage, we aim to forecast the future $H$ steps $\left[p_{T+1}, \cdots, p_{T+H}\right]$. For simplicity, we denote this target $H$-step series as $y_{T}$. The forecasting model takes the previous $M$ steps $\left[p_{T-M+1}, \cdots, p_{T-1}, p_{T}\right]$ as input, simplified as $x_{T}$. On certain tasks, additional content information, denoted as $c_{T}$, may also be provided. The forecasting model $M(\cdot)$ then predicts the target as follows:
\begin{equation}
\label{eq:forecast}
\begin{aligned}
    \hat{y}_{T} &= M(x_{T}, c_{T}), \\
    \hat{y}_{T} &\triangleq \left[\hat{p}_{T+1}, \cdots, \hat{p}_{T+H}\right], \\
    x_{T} &\triangleq \left[p_{T-M+1}, \cdots, p_{T-1}, p_{T}\right],
\end{aligned}
\end{equation}
where $\hat{y}_{T}$ denotes the predicted result. The training objective of the forecasting model is to reduce the distance between the forecasted value $\hat{y}_{T}$ and the ground truth $y_{T}$. A certain prediction loss $\mathcal{L}_p$ is adopted to measure such distances. Hence, the training objective of the forecasting model can be denoted as follows:
\begin{equation}
\label{eq:loss_forecast}
    \mathcal{L}_p = \frac{1}{|\mathcal{D}|} \min_{M(\cdot)} \sum_{(x_T, y_T, c_T) \in \mathcal{D}} \mathcal{l}_p(y_{T}, \hat{y}_{T}).
\end{equation}
Note that the Mean-Square-Error (MSE) Loss is widely adopted as the prediction loss $\mathcal{l}_p(\cdot)$ on each data instance~\cite{PatchTST,DLinear}. 

\subsubsection{Optimization Stage}
After obtaining the prediction result $\hat{y}_T$ from the well-trained forecasting model $M(\cdot)$, Equation (\ref{eq:formulation}) can be viewed as an optimization problem via replacing the parameters $y_T$ with the prediction result $\hat{y}_T$. Hence, Equation (\ref{eq:formulation}) is derived into:

\begin{equation}
\label{eq:pto_optimize}
\begin{aligned}
    & \min \ \mathbf{a} \cdot \hat{y}_T \\
    & s.t. \sum \mathbf{a} = 1, \mathbf{a} \in \mathcal{A}, \mathcal{A} \subseteq [0, 1]^{H}.
\end{aligned}
\end{equation}

The above equation can be observed as optimization $\mathbf{a}$ while treating the forecasted result $\hat{y}_T$ as ground truth values. Hence, the accuracy of $\hat{y}_T$ becomes important for the quality of the final decision. However, it is recognized that accurately forecasting time series is not an easy task~\cite{neurips24talk}. To reduce the side effects of inaccurate prediction, we additionally propose forecasting uncertainty constraints that adaptively adjust themselves to constraints on the feasible position of allocation. Suppose a risk measurement can be obtained on each allocation position. As we will elaborate in Section~\ref{sec:method_uncertainty}, the risk vector $\mathbf{r}$ can be represented as:
\begin{equation}
\label{eq:risk_space}
    \mathbf{r} \in \mathbb{R}_{\ge 0}^{H}, \ \mathbb{R}_{\ge 0} = \{x \in \mathbb{R} \ | \ x \ge 0 \}
\end{equation}
Hence, we define a new risk-aware feasible space $\mathcal{A}^{'}(\mathbf{r})$ as follows:
\begin{equation}
    \mathcal{A}^{'}(\mathbf{r}) = \{ \mathbf{a} \cdot \mathbf{r} \le r_0 \ | \ \mathbf{a} \in \mathcal{A}\}
\end{equation}  
Here, $r_0$ is a pre-defined scalar representing the risk tolerance level. A smaller $r_0$ would lead to a tighter constraint on the forecasting uncertainty and a stronger preference towards forecasting results with high confidence. It is easy to observe that $\mathcal{A}^{'}(\mathbf{r}) \subseteq \mathcal{A}$. Correspondingly, the objective of Equation (\ref{eq:pto_optimize}) becomes to solve the following task: 
\begin{equation}
\label{eq:pto_optimize2}
\begin{aligned}
    & \mathbf{a}^*(\hat{y}_T) = \argmin \ \mathbf{a}(\hat{y}_T) \cdot \hat{y}_T \\
    & s.t. \sum \mathbf{a}(\hat{y}_T) = 1, \mathbf{a}(\hat{y}_T) \in \mathcal{A}^{'}(\mathbf{r)}, \mathcal{A}^{'}(\mathbf{r}) \subseteq [0, 1]^{H}.
\end{aligned}
\end{equation}
Here $\mathbf{a}^*(\hat{y}_T)$ refers to the optimal allocation under the prediction $\hat{y}_T$. It is also known as the prescriptive decision~\cite{Prescriptive}.
Once the future asset price $y_T$ is known, the optimal allocation results in $\mathbf{a}^{*}(y_T)$ can be readily obtained by solving Equation (\ref{eq:formulation}) as a continuous optimization problem. $\mathbf{a}^{*}(y_T)$ is also known as the full-information optimal decision~\cite{Prescriptive}.

After obtaining both the optimal allocation $a^*(y_T)$ and prescriptive decision $a^*(\hat{y}_T)$, we can use the \textit{regret} metric, defined as the cost gap between these two allocation plans, to evaluate the quality of the decision. This can be written as:
\begin{equation}
\label{eq:regret}
    \text{regret} \triangleq |\mathbf{a}^{*}(y_{T}) \cdot y_{T} - \mathbf{a}^*(\hat{y}_{T}) \cdot y_{T}|.
\end{equation}
A lower regret indicates the predicted allocation $\mathbf{a}^*(\hat{y}_T)$ is closer to the optimal allocation $\mathbf{a}^*(y_T)$, indicating a lower operation cost and a better decision quality.

\subsubsection{Uncertainty Quantify via Conformal Prediction}
\label{sec:method_uncertainty}

In this section, we focus on how to quantify the uncertainty of an arbitrary forecasting model and, more importantly, how to utilize such uncertainty to guide the training of our framework. Motivated by previous work~\cite{CF-RNN}, we adopt conformal prediction to measure the positional uncertainty of time series forecasting. The pseudo-code for positional uncertainty calculation is shown in Algorithm~\ref{alg:uncertainty}.

\begin{algorithm}
\caption{Calculating Positional Uncertainty for Backbone}

\label{alg:uncertainty}
\begin{algorithmic}[1]

    \Require{Calibration Dataset $\mathcal{D}_c$, coverage rate $\gamma$}
    \Ensure{Positional Uncertainty $\mathbf{r}$}

    \State{Initialize Positional Uncertainty Sets $\epsilon_1$ = \{\ \}, $\cdots$, $\epsilon_H$ = \{\ \}}

    \For{for data instance $(x_T, y_T, c_T)$ in Calibration Set $\mathcal{D}_c$}
    \State{Calculate $\hat{y}_T = \left[\hat{p}_{T+1}, \cdots \hat{p}_{T+H}\right]$ given Eq.~\ref{eq:forecast}}
    \For{$h$ in $1, \cdots, H$}
        \State{$\epsilon_h \leftarrow \epsilon_h \cup \{|\hat{p}_{T+h} - p_{T+h}|\}$}
    \EndFor
    \EndFor

    \For{$h$ in $1, \cdots, H$}
    \State{$r_h = \left(\frac{|\mathcal{D}_c| + 1}{|\mathcal{D}_c|} \gamma\right)$ - quantile in $\epsilon_h$}
    \EndFor

    \State{Return $\mathbf{r} = [r_1, r_2, \cdots, r_H]$}
\end{algorithmic}
\end{algorithm}

\subsubsection{Overall Training Process of RTS-PtO}
The pseudo-code for the training of the RTS-PtO framework is shown in Algorithm~\ref{alg:train_pto}.

\begin{algorithm}
\caption{The Training Process of RTS-PtO Framework}

\label{alg:train_pto}
\begin{algorithmic}[1]

    \Require{Dataset $\mathcal{D}$, risk tolerance $r_0$, epoch number $T$}
    \Ensure{A allocation function $\mathbf{a}^*(\hat{y}_T)$ produce allocation with forecasting result $\hat{y}_T$}

    \For{Epoch t = 1, $\cdots$, $T$}
    \State{Update the forecasting model $M$ given Eq.~\ref{eq:final}}
    \EndFor

    \State{Obtain the positional uncertainty $\mathbf{r}$ for epoch $t$ given Alg.~\ref{alg:uncertainty}} 

    \State{Obtain the allocation feasiable space $\mathcal{A}(\mathbf{r})$ given Eq.~\ref{eq:risk_space}}

    \State{Obtain the prescriptive decision $\mathbf{a}^*(\hat{y}_T)$ given Eq.~\ref{eq:pto_optimize2}}

\end{algorithmic}
\end{algorithm}

\subsection{RTS-PnO: An End-to-end Solution with Adaptive Uncertainty Constraints}
\label{sec:method_rtspno}

In this section, we propose our model-agnostic framework, RTS-PnO. As stated in Section~\ref{sec:introduction}, the model differs from the previous two-stage solution in two aspects: (i) an end-to-end training predict-and-optimize paradigm aiming at aligning both the training objective and business goal and (ii) an adaptive risk-aware constraint to mitigate the forecasting error of the prediction model. We will discuss them in the following paragraphs.

\subsubsection{End-to-end training with PnO framework}
Unlike the PtO framework, the predict-and-optimize (PnO) framework directly trains the model with feedback from the optimization stage. To align both the training process and optimization goal, a surrogate loss is proposed to make the optimization process in the second stage differentiable, denoted as:
\begin{equation}
    \mathcal{L}_o = \frac{1}{|\mathcal{D}|} \min_{\mathbf{M}(\cdot)} \sum_{(\hat{a}(\hat{y}_T), \mathbf{a}(y_T)) \in \mathcal{D}} \mathcal{l}_o (\mathbf{a}^*(\hat{y}_T), \mathbf{a}(y_T))
\end{equation}
Then also exist several opinions for the surrogate loss $\mathcal{L}_{o}(\cdot)$ for the optimization stage. Here, we adopt the SPO+ loss~\cite{SPO+}, which is widely adopted in classic operation research problems and is verified to have outstanding performances~\cite{PTO-PNO-Benchmark,DFL-Survey}. The SPO+ loss is optimized directly on the prescriptive decision $\mathbf{a}^*(\hat{y}_T)$, denoted as:
\begin{equation}
\label{eq:loss_spo+}
    \mathcal{l}_o (\mathbf{a}^*(\hat{y}_T)) \triangleq 2 \mathbf{a}^*(y_T) \hat{y}_T - \mathbf{a}^*(y_T) y_T + \max_{\mathbf{a} \in \mathcal{A}} \{\mathbf{a} y_T - 2 \mathbf{a} \hat{y}_t \}.
\end{equation}

\begin{figure}[!htbp]
    \centering
    \includegraphics[width=0.49\textwidth]{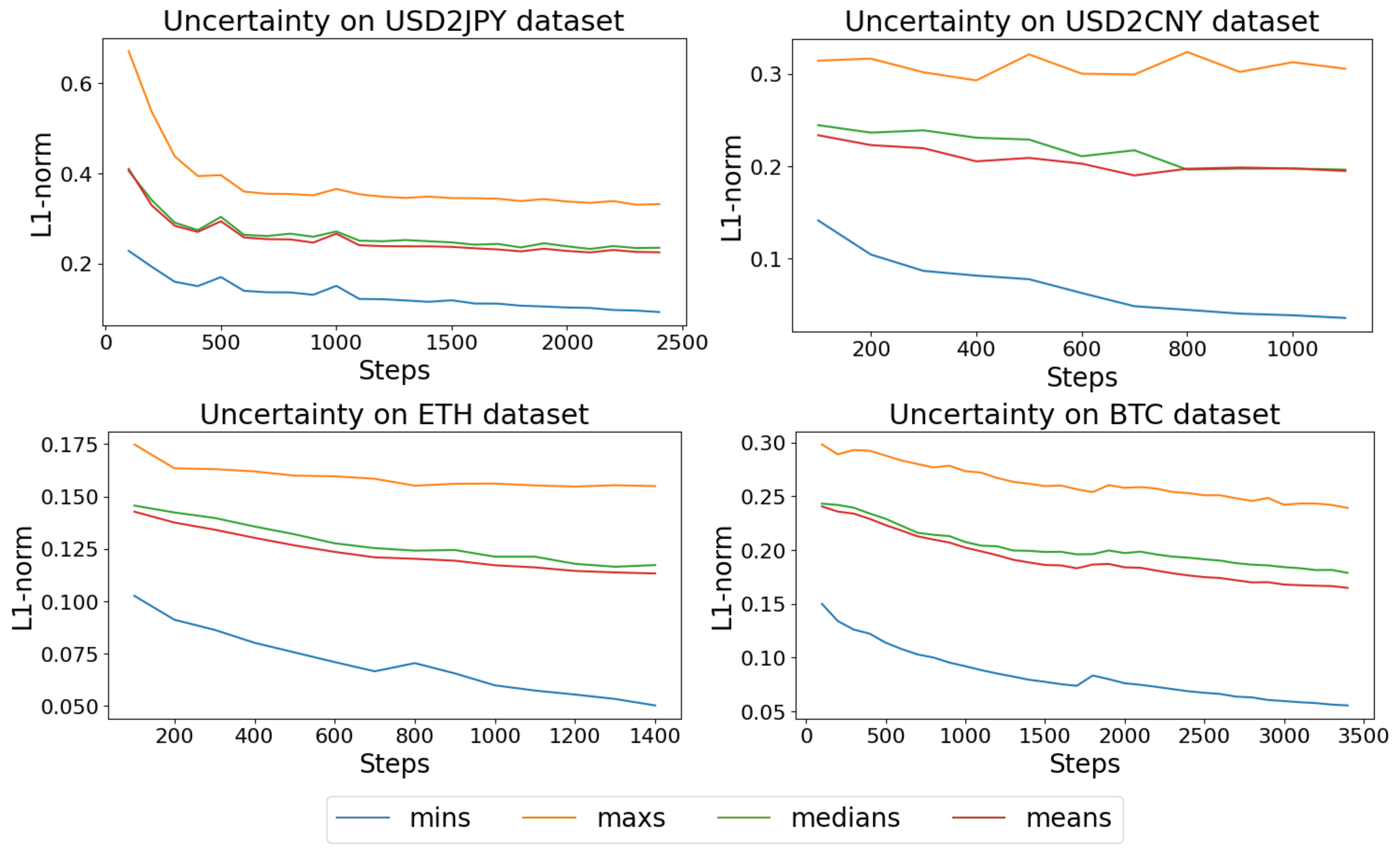}
     \caption{Empirical observation on the trending of uncertainty during the training process. Here, we report the minimum, maximum, mean, and median of the positional uncertainty.}
    \Description{Empirical Observation on the Trending of Uncertainty during the Training Process.}
    \label{fig:empirical}
\end{figure}

\subsubsection{Adaptive Uncertainty Constraints}
\label{sec:method_uncertainty_adaptive}
Although the SPO+ loss~\cite{SPO+} theoretically aligns the training objective and business objective, empirical experiments suggest there is still a significant gap between these two~\cite{PTO-PNO-Benchmark}. Hence, it is deemed necessary to introduce the uncertainty constraint as a mitigation towards inaccurate prediction and objective mismatch. 

As the risk vector $\mathbf{r}$ reflects the forecasting uncertainty given the position, it is easy to notice that as the parameters within the forecasting model update, its forecasting uncertainty changes accordingly. The empirical studies in Figure~\ref{fig:empirical} on four datasets show that the uncertainty gradually reduces during the training process. 
Therefore, adopting a fixed constraint like the PtO framework would yield an outdated uncertainty constraint and a suboptimal decision. 
To solve such a problem, we propose to add an adaptive risk constraint that updates the risk vector $\mathbf{r}_t$ simultaneously after each epoch following Algorithm~\ref{alg:uncertainty}. Such a design would better reflect the current uncertainty of the forecasting model. Additionally, it becomes hard to fix a constant on the risk-tolerance threshold $r_0$. Therefore, we define $r_0$ as the $\alpha$-quantile of the risk vector $\mathbf{r}$, denoted as:
\begin{equation} \label{eq:risk_r0}
    r_0 = \alpha-\text{quantile in } \mathbf{r}
\end{equation}
Hence, we define a new risk-aware feasible space $\mathcal{A}^{'}(\mathbf{r}_t)$ as follows:
\begin{equation} 
    \mathcal{A}^{'}(\mathbf{r}_t) = \{ \mathbf{a} \cdot \mathbf{r}_t \le \alpha-\text{quantile in } \mathbf{r}_t \ | \ \mathbf{a} \in \mathcal{A}\}
\end{equation}
Correspondingly, Equation (\ref{eq:loss_spo+}) derives into:
\begin{equation}
    \mathcal{l}_o (\mathbf{a}^*(\hat{y}_T), \mathbf{r}) \triangleq 2 \mathbf{a}^*(y_T) \hat{y}_T - \mathbf{a}^*(y_T) y_T + \max_{\mathbf{a} \in \mathcal{A}(\mathbf{r}_t)} \{\mathbf{a} y_T - 2 \mathbf{a} \hat{y}_T \}.
\end{equation}
The final training objective of RTS-PnO can be written as:
\begin{equation} \label{eq:final}
    \min_{\mathbf{M}(\cdot)} \mathcal{L}_o + \beta \cdot \mathcal{L}_p,
\end{equation}
where the prediction loss $\mathcal{L}_p$ is introduced as a regulator to balance the tradeoff between forecasting accuracy and decision quality, and $\beta$ denotes its coefficient.

\subsubsection{Overall Training Process of RTS-PnO}
The pseudo-code for the training of the RTS-PnO framework is shown in Algorithm~\ref{alg:train}.

\begin{algorithm}
\caption{The Training Process of RTS-PnO Framework}

\label{alg:train}
\begin{algorithmic}[1]

    \Require{Dataset $\mathcal{D}$, uncertainty quantile $\alpha$, loss balancer $\beta$, epoch number $T$}
    \Ensure{A allocation function $\mathbf{a}^*(\hat{y}_T)$ produce allocation with forecasting result $\hat{y}_T$, a forecasting model $M(\cdot)$ outputs forecasting result $\hat{y}_T$}

    \For{Epoch t = 1, $\cdots$, $T$}

    \State{Update the forecasting model $M$ given Eq.~\ref{eq:final}}

    \State{Update the positional uncertainty $\mathbf{r}_t$ for epoch $t$ given Alg.~\ref{alg:uncertainty}} 

    \State{Update the allocation feasible space $\mathcal{A}(\mathbf{r}_t)$ given Eq.~\ref{eq:risk_space}}
    \EndFor

\end{algorithmic}
\end{algorithm}

\section{Experiment}
\label{sec:experiment}

In this section, to comprehensively evaluate our proposed RTS-PnO, we design experiments to answer the following research questions: 

\begin{itemize}[topsep=0pt,noitemsep,nolistsep,leftmargin=*]
    \item \textbf{RQ1}: Can RTS-PnO achieve superior performance in terms of decision quality compared with other baselines?
    \item \textbf{RQ2}: How does the forecasting model influence the overall performance in terms of decision quality?
    \item \textbf{RQ3}: How does the adaptive uncertainty design influence the performance in terms of decision quality?
    \item \textbf{RQ4}: How efficient is RTS-PnO compared to other methods?
    \item \textbf{RQ5}: How does the Predict-and-Optimize design influence the forecasting performance?
\end{itemize}
\begin{table*}[!htbp]
    \centering
    \caption{Main Experiment with PatchTST as Forecasting Model}
    \label{tab:main}
    \resizebox{0.98\textwidth}{!}{
    \begin{tabular}{c|c|cccc|cccc|cc|cc|cc}
    \hline
        \multirow{3}{*}{Category} & \multirow{3}{*}{Dataset} & \multicolumn{4}{c|}{Forecasting-Only} & \multicolumn{4}{c|}{Risk-Avoid} & \multicolumn{2}{c|}{\multirow{2}{*}{RTS-PtO}} & \multicolumn{2}{c|}{\multirow{2}{*}{RTS-PnO}} & \multicolumn{2}{c}{Relative} \\
    \cline{3-10}
        & & \multicolumn{2}{c|}{Top-1} & \multicolumn{2}{c|}{Top-5} & \multicolumn{2}{c|}{Top-1} & \multicolumn{2}{c|}{Top-5} & \multicolumn{2}{c|}{} & \multicolumn{2}{c|}{} & \multicolumn{2}{c}{Improvement} \\
    \cline{3-16}
        & & regret$\downarrow$ & R.R.$\downarrow$ & regret$\downarrow$ & R.R.$\downarrow$ & regret$\downarrow$ & R.R.$\downarrow$ & regret$\downarrow$ & R.R.$\downarrow$ & regret$\downarrow$ & R.R.$\downarrow$ & regret$\downarrow$ & R.R.$\downarrow$ & regret(\%) & R.R.(\%) \\
    \hline
        \multirow{4}{*}{Currency}  
        & USD2CNY & 36.88 & 5.10  & 37.00 & 5.12 & 35.80 & 4.95 & 35.83 & 4.96 & \underline{35.74} & \underline{4.94} & \textbf{31.68} & \textbf{4.38} & 12.82\% & 12.79\% \\
        & USD2JPY & 54.50 & 34.92 & 54.21 & 34.73 & \underline{49.66} & \underline{31.90} & 50.01 & 32.12 & 52.11 & 32.66 & \textbf{48.77} & \textbf{31.25} & 1.82\% & 2.08\% \\
        & AUD2USD & 19.56 & 29.60 & 19.92 & 30.15 & \underline{19.38} & \underline{29.36} & 19.49 & 29.52 & 19.48 & 29.51 & \textbf{19.06} & \textbf{28.84} & 1.68\% & 1.80\% \\
        & NZD2USD & 17.43 & 28.75 & 17.66 & 29.14 & \underline{16.54} & \underline{27.29} & 16.64 & 27.44 & 16.82 & 27.75 & \textbf{15.68} & \textbf{25.85} & 5.48\% & 5.57\% \\
    \hline
        \multirow{2}{*}{Stock}
        & S\&P 500  & 134.99  & 4.25 & 135.47 & 4.24 & \textbf{122.50} & \textbf{3.84} & 124.24 & \underline{3.90} & 126.06  & 3.94 & \underline{124.05} & \underline{3.90} & -1.27\% & -1.56\% \\
        & Dow Jones & 1090.88 & 4.16 & 1075.79 & 4.09 & \underline{1022.73} & \underline{3.91} & 1032.21 & 3.93 &  1022.90 & 3.92 & \textbf{997.52} & \textbf{3.82} & 2.53\% & 2.36\% \\
    \hline
        \multirow{2}{*}{Cryptos} 
        & BTC & 2159.78 & 4.46 & 2167.96 & 4.47 & \underline{1856.21} & \underline{3.90} & 1858.57 & 3.91 & 1924.65 & 3.96 & \textbf{1843.26} & \textbf{3.70} & 0.70\% & 5.41\% \\
        & ETH & 151.14  & 5.56 & 149.61 & 5.48 & \underline{131.41} & \textbf{4.68} & 131.42 & \textbf{4.68} & 138.60 & 4.96 & \textbf{131.40} & \underline{4.73} & 0.00\% & -1.07\% \\
    \hline
        \multicolumn{2}{c|}{Avg. Rank} & 5.38 & 5.5 & 5.63 & 5.5 & 2 & 1.88 & 3.38 & 3.13 & 3.5 & 3.5 & 1.13 & 1.25 \\
    \hline
    \end{tabular}}
\begin{tablenotes}
\footnotesize
\item[1] Here \textbf{bold} font indicates the best-performed method and \underline{underline} font indicates the second best-performed method. We also report the average rank of each method across all datasets. Notice that for the Currency category, all \textit{regret} and \textit{R.R.} omit the scaler $\times 10^{-4}$. For the rest category, all \textit{R.R.} omit the scaler $\times 10^{-2}$.
\end{tablenotes}
\end{table*}

\subsection{Experimental Setup}

\subsubsection{Time-Series Forecasting Models as Backbones}

In the experiment, we adopt four SOTA time-series forecasting models as the backbone. PatchTST~\cite{PatchTST} is adopted as the default backbone without specification. DLinear~\cite{DLinear}, TimesNet~\cite{TimesNet} and FEDformer~\cite{FEDformer} are also included as backbones. 
Below, we include a brief explanation of each backbone involved during the experiments.
\begin{itemize}[topsep=0pt,noitemsep,nolistsep,leftmargin=*]
    \item PatchTST~\cite{PatchTST}: An encoder-only Transformer that operates on patches instead of individual time steps to quadratically reduce computational costs and model patch-wise dependencies.
    \item DLinear~\cite{DLinear}: A fully linear model with a decomposer to separate seasonal and trend signals. The two signals are processed by separate linear transformations and added before output.
    \item TimesNet~\cite{TimesNet}: A modern temporal convolutional architecture that reshapes the 1D series into 2D matrices by salient periodicities in each model block to jointly capture intra-period and inter-period dependencies.
    \item FEDformer~\cite{FEDformer}: An encoder-decoder Transformer that computes attention in the Fourier space after filtering.
\end{itemize}
We noticed that some of the models (TimesNet~\cite{TimesNet} and FEDformer~\cite{FEDformer}) make use of timestamp information by incorporating special temporal embeddings, whereas other models choose not to. We remove the time stamp information from all models to ensure fair comparisons.

\subsubsection{Baselines}

In the following experiment, we adopt the following optimization baselines:
\begin{itemize}[topsep=0pt,noitemsep,nolistsep,leftmargin=*]
    \item Forecasting-Only: The forecasting-only method only makes the decision based on the forecasted result. Specifically, it follows a greedy approach by selecting the lowest k points in the future and evenly distributing the asset quota among these time steps. Here, we adopt Top-1 and Top-5 as baselines.
    \item Risk-Avoiding: The Risk-Avoiding method focuses on the worst-case scenarios. Specifically, it selects the k points with the lowest upper bound for the forecasting and evenly distributes the asset quota among these time steps. It adopts the uncertainty quantification approach in Algorithm~\ref{alg:uncertainty} to calculate the upper bound. Here, we adopt Top-1 and Top-5 as baselines.
\end{itemize}

\subsubsection{Benchmarks}

We evaluate the performance on eight datasets: USD2CNY, USD2JPY, AUD2USD, NZD2USD, S\&P 500, Dow Jones, BTC, ETC, from three financial scenarios: Currency, Stock, and Crypto.
The statistics of these datasets are shown in Table~\ref{tab:statistics}. The processed dataset is available here\footnote{https://github.com/fuyuanlyu/RTS-PnO}. 
Below we further describe the details of each dataset.
\begin{itemize}[topsep=0pt,noitemsep,nolistsep,leftmargin=*]
    \item Currency: The Currency datasets, USD2CNY, USD2JPY, AUD2USD and NZD2USD, contain 10-minute currency between different currency pairs. The date of this dataset ranges from 2023/07/10 to 2024/07/08. We removed the intervals where the market was closed. Here USD represents the US Dollar, CNY represents the Chinese Yuan, JPY represents the Japanese Yen, AUD represents the Australia Dollar, and NZD represents the New Zeland Dollar.
    \item Stock~\cite{dataset-Stock}: The Stock datasets, S\&P 500 and Dow Jowes, contain daily prices from 1990/01/03 to 2024/2/16. It only contains workdays within this period. We use the Open market price as default.
    \item Cryptos~\cite{dataset-Crypto}: The original Cryptos data is sampled at a minute frequency. We observed that the series remains relatively stable within each hour, and the first half of the data exhibits minimal variation. Therefore, we retain only the second half of the data and downsample it to an hourly frequency. We use the Open market price as default. Eventually, the ETH dataset contains hourly prices from 2020/07/18 to 2024/07/28, while the BTC dataset contains hourly prices from 2019/11/27 to 2024/07/29.
\end{itemize}

\begin{table}[!htbp]
    \centering
    \caption{Benchmark Statistics}
    \label{tab:statistics}
    \resizebox{0.48\textwidth}{!}{
    \begin{tabular}{c|c|c|cccc}
    \hline
        Category & Dataset & \#Times & Max & Min & Mean & Median \\
    \hline
        \multirow{4}{*}{Currency}
        & USD2CNY & \multirow{4}{*}{22968} 
                    & 7.0905 & 7.3494 & 7.2267 & 7.2333 \\
        & USD2JPY & & 1.3740 & 1.6195 & 1.4949 & 1.4922 \\
        & AUD2USD & & 0.6274 & 0.6895 & 0.6559 & 0.6562 \\
        & NZD2USD & & 0.5774 & 0.6411 & 0.6066 & 0.6087 \\
    \hline
        \multirow{2}{*}{Stock}
        & S\&P 500 & \multirow{2}{*}{8597} 
                      & 295.46 & 5029.73 & 1596.80 & 1270.20 \\
        & Dow Jones & & 2365.10 & 38797.90 & 13663.80 & 10846.30 \\
    \hline
        \multirow{2}{*}{Cryptos}
        & BTC & 40932 & 4206.86 & 73705.36 & 32269.79 & 29352.15 \\
        & ETH & 35301 & 233.72 & 4853.69 & 2126.29 & 1886.80 \\
    \hline
    \end{tabular}}
\end{table}

\subsubsection{Metrics}
To measure the decision quality of different methods, we adopt the commonly used metric \textit{regret} defined in Equation (\ref{eq:regret}) as the metric~\cite{DFL-Survey,PTO-PNO-Benchmark,SPO+}. Considering the rapid changing of the asset price in certain datasets, such as BTC and ETH from the Cryptos domain, solely adopting \textit{regret} would favor the model making a good decision in extreme cases. Hence, we additionally adopt the \textit{relative regret}, abbreviated as \textit{R-R}, which denotes the \textit{regret} with the optimal value at that time step. This can be formulated as:
\begin{equation} \label{eq:rel_regret}
    \text{R.R.} \triangleq \frac{\text{regret}}{\text{optimal cost}} = \frac{|\mathbf{a}^*(y_T) \cdot y_T - \mathbf{a}^*(\hat{y}_T) \cdot y_T|}{\mathbf{a}^*(y_T) \cdot y_T}.
\end{equation}
For both \textit{regret} and \textit{R.R.}, a lower value indicates the decision is closer to the optimal one, which is naturally a higher-quality decision.

Apart from the decision quality, we also need to measure the forecasting ability in RQ5. Here, we adopt \textit{MSE} and \textit{MAE} as forecasting metrics, following previous works in the time-series forecasting domain~\cite{PatchTST,TimesNet,DLinear,FEDformer}. For both \textit{MSE} and \textit{MAE}, a lower value indicates a higher forecasting accuracy.

\subsubsection{Implementation Details} In this section, we provide the implementation details for all offline experiments. Our implementation is based on the PyTorch framework. Adam Optimizer is adopted for all setups. We select the learning ratio from \{1e-3, 3e-4, 1e-4, 3e-5, 1e-5\}. 
We adopt Gurobi~\cite{Gurobi} as the solver for optimization problems and borrow the implementation of SPO+ loss~\cite{SPO+} from the PyEPO~\cite{PyEPO} library.
All experiments in this section are run on an Nvidia RTX 4090D (24GB) GPU with 8 Intel (R) Xeon (R) Platinum 8481C and 40GB of memory.

\begin{table*}[!htbp]
    \centering
    \caption{Ablation Study on Forecasting Models}
    \label{tab:ablation_model}
    \resizebox{0.98\textwidth}{!}{
    \begin{tabular}{c|c|cccc|cccc|cc|cc|cc}
    \hline
        \multirow{2}{*}{Forecasting} & \multirow{3}{*}{Dataset} & \multicolumn{4}{c|}{Forecasting-Only} & \multicolumn{4}{c|}{Risk-Avoiding} & \multicolumn{2}{c|}{\multirow{2}{*}{RTS-PtO}} & \multicolumn{2}{c|}{\multirow{2}{*}{RTS-PnO}} & \multicolumn{2}{c}{Relative} \\
    \cline{3-10}
        \multirow{2}{*}{Model} & & \multicolumn{2}{c|}{Top-1} & \multicolumn{2}{c|}{Top-5} & \multicolumn{2}{c|}{Top-1} & \multicolumn{2}{c|}{Top-5} & \multicolumn{2}{c|}{ } & \multicolumn{2}{c|}{ } & \multicolumn{2}{c}{Improvement} \\
    \cline{3-16}
        & & regret$\downarrow$ & R.R.$\downarrow$ & regret$\downarrow$ & R.R.$\downarrow$ & regret$\downarrow$ & R.R.$\downarrow$ & regret$\downarrow$ & R.R.$\downarrow$ & regret$\downarrow$ & R.R.$\downarrow$ & regret$\downarrow$ & R.R.$\downarrow$ & regret(\%) & R.R.(\%) \\
    \hline
        \multirow{2}{*}{DLinear}
        & USD2CNY   & 36.99 & 5.12 & 36.73 & 5.08 & 35.50 & \underline{4.91} & 38.11 & 5.27 & \underline{35.31} & 4.98 & \textbf{34.88} & \textbf{4.81} & 1.23\% & 3.50\% \\ 
        & Dow Jones & 1103.11 & 4.21 & 1128.71 & 4.24 & \textbf{1036.65} & \textbf{3.96} & 1075.97 & 4.08 & 1073.30 & 4.10 & \underline{1042.35} & \underline{3.98} & -0.55\% & -0.51\% \\
    \hline
        \multirow{2}{*}{TimesNet}
        & USD2CNY   & 39.77 & 5.50 & 39.46 & 5.46 & 36.83 & 5.09 & 37.47 & 5.18 & \underline{35.99} & \underline{4.98} & \textbf{33.73} & \textbf{4.66} & 6.70\% & 6.87\% \\
        & Dow Jones & 1157.76 & 4.40 & 1143.82 & 4.32 & \underline{1037.71} & 3.98 & 1082.45 & 4.11 & 1042.67 & \underline{3.95} & \textbf{972.51} & \textbf{3.74} & 6.70\% & 5.61\% \\
    \hline
        \multirow{2}{*}{FEDFormer}
        & USD2CNY   & 36.44 & 5.04 & 36.89 & 5.10 & 36.28 & 5.02 & 36.53 & 5.05 & \underline{35.94} & \underline{4.97} & \textbf{32.32} & \textbf{4.47} & 11.23\% & 11.19\% \\
        & Dow Jones & 1087.49 & 4.15 & 1100.99 & 4.19 & 1065.08 & 4.05 & 1078.61 & 4.09 & \underline{1043.41} & \underline{3.98} & \textbf{1010.96} & \textbf{3.82} & 3.21\% & 4.19\% \\
    \hline
    \end{tabular}}
\begin{tablenotes}
\footnotesize
\item[1] Here \textbf{bold} font indicates the best-performed method and \underline{underline} font indicates the second best-performed method. Notice that for the USD2CNY dataset, all \textit{regret} and \textit{R.R.} omit the scaler $\times 10^{-4}$. For the Dow Jones dataset, all \textit{R.R.} omit the scaler $\times 10^{-2}$.
\end{tablenotes}
\end{table*}

\subsection{Main Experiment (RQ1)}

The overall performance of the proposed RTS-PnO and other baselines on eight benchmarks is reported in Table~\ref{tab:main}. We summarize our observation below.

First, our RTS-PnO proves to be effective compared with all other baselines in terms of both absolute regret and relative regret. In most datasets, the RTS-PnO method demonstrated improvements in terms of absolute regret (Abs-R) and relative regret (Rel-R). The RTS-PnO framework also achieved the best average ranking in both absolute and relative risks. However, the improvement brought by RTS-PnO differs on various datasets. For instance, on the USD2CNY dataset, RTS-PnO brings 12.82\% and 12.79\% improvement in terms of absolute and relative regrets. In contrast, on the S\&P 500 dataset, RTS-PnO ranks second among all baselines.

Secondly, the RTS-PtO framework outperforms the forecasting-only approach in multiple datasets, especially in the stock market (e.g., S\&P 500 and Dow Jones). This indicates that combining prediction and optimization in a two-stage process can improve decision quality.

Thirdly, the risk-avoiding strategy performed well across multiple datasets, particularly in the cryptocurrency and stock domains. For instance, on both S\&P 500 and ETH datasets, risk-avoiding strategies rank first and show reductions in absolute and relative regrets, suggesting their effectiveness in highly volatile markets. Moreover, in all cases, the risk-avoiding strategy outperforms its forecasting-based versions, showing its adaptiveness in various cases.

Finally, the advantage of adopting a Top-k decision instead of a Top-1 decision varies in Forecasting-Only and Risk-Avoiding scenarios. In all cases, the Top-1 version outperforms the Top-5 version in the Risk-Avoiding scenario, indicating the conflict between the heuristic risk-avoiding strategy and the uncertainty qualification approach. However, in the Forecasting scenario, the Top-1 and Top-5 strategies vary on different datasets, showing their limitations in making decisions under rapidly changing markets.

\subsection{Ablation Study on Forecasting Models (RQ2)}

The RTS-PnO framework is proposed as a model-agnostic framework to adopt the rapidly changing time-series forecasting models as baselines seamlessly. This section showcases its compatibility with three different models: DLinear, TimesNet, and FEDformer, each representing one category for the time-series forecasting model. The experiment is conducted on two datasets: USD2CNY as a representative for currency data and Dow Jones as a representative for stock data. The result is shown in Table~\ref{tab:ablation_model}. Based on the results, we can make the following observations:

First, we observe consistent performance improvement like in the previous section. The RTS-PnO framework demonstrates significant performance improvements across various forecasting models, e.g., DLinear, TimesNet, and FEDFormer. For instance, in the USD2CNY dataset, the FEDFormer model achieved improvements of 11.23\% and 11.19\% in absolute regret (Abs-R) and relative regret (Rel-R), respectively, when using RTS-PnO. This indicates that the RTS-PnO framework is not dependent on a specific model.

Secondly, the selection of the forecasting model also influences the decision quality. For instance, TimesNet yields the best performance on the Dow Jones dataset. Meanwhile, PatchTST outperforms others on USD2CNY datasets. Such an observation further highlights the importance of a model-agnostic framework. The compatibility of such a framework greatly extends its application in various real-world applications.

Thirdly, we also observe that the RTS-PtO framework and risk-avoiding paradigm are competitive approaches across various forecasting models. For instance, Risk-Avoiding with Top-1 decision yields best on Dow Jones with the DLinear model, while the RTS-PtO framework ranks second in four out of six cases.

\subsection{Ablation Study on Adaptive Uncertainty Constraint for RTS-PnO (RQ3)}

To investigate the effect of the adaptive uncertainty constraint on the decision quality, we conduct the following ablation study. In this study, we replace the adaptive uncertainty constraint in Section~\ref{sec:method_uncertainty_adaptive} with the fixed uncertainty constraint in Section~\ref{sec:method_uncertainty} and Equation (\ref{eq:risk_space}). The evaluation is conducted over six datasets, shown in Table~\ref{tab:ablation_uncertainty}.

\begin{table}[!htbp]
    \centering
    \caption{Ablation on Uncertainty Constraint}
    \label{tab:ablation_uncertainty}
    \resizebox{.48\textwidth}{!}{
    \begin{tabular}{c|cc|cc|cc}
    \hline
        \multirow{2}{*}{Dataset} & \multicolumn{2}{c|}{PtO} & \multicolumn{2}{c|}{Fixed-PnO} & \multicolumn{2}{c}{Adaptive-PnO} \\
    \cline{2-7}
        & regret$\downarrow$ & R.R.$\downarrow$ & regret$\downarrow$ & R.R.$\downarrow$ & regret$\downarrow$ & R.R.$\downarrow$ \\
    \hline
        USD2CNY & 35.74 & 4.94 & \underline{34.66} & \underline{4.71} & \textbf{31.68} & \textbf{4.38} \\
        USD2JPY & 52.11 & 32.66 & \underline{49.21} & \underline{31.83} & \textbf{48.77} & \textbf{31.25} \\
        S\&P 500  & \underline{126.06} & \underline{3.94} & 129.13 & 4.02 & \textbf{124.05} & \textbf{3.90} \\
        Dow Jones & \underline{1022.90} & \underline{3.92} & 1026.45 & 3.92 & \textbf{997.52} & \textbf{3.82} \\
        BTC & \underline{1924.65} & \underline{3.96} & 1939.81 & 4.02 & \textbf{1843.26} & \textbf{3.70} \\
        ETH & 138.60 & 4.96 & \underline{136.66} & \underline{4.90} & \textbf{131.40} & \textbf{4.73} \\
    \hline
    \end{tabular}}
\begin{tablenotes}
\footnotesize
\item[1] Here \textbf{bold} font indicates the best-performed method and \underline{underline} font indicates the second best-performed method. Notice that for the USD2CNY and USD2JPY dataset, all \textit{regret} and \textit{R.R.} omit the scaler $\times 10^{-4}$. For the rest datasets, all \textit{R.R.} omit the scaler $\times 10^{-2}$.
\end{tablenotes}
\end{table}

We can make the following observations:
First, we can easily observe that the PnO framework with adaptive uncertainty quantification constantly outperforms the fixed one. 
Secondly, it is easy to observe that PnO with fixed uncertainty is sometimes outperformed by its PtO version. Specificially, PtO ranks second on S\&P 500, Dow Jones, and BTC datasets, while PnO with fixed uncertainty ranks second on the rest datasets. This is likely because the fixed uncertainty, reflecting the uncertainty of a well-trained forecasting model, misleads the PnO framework during the training phase, particularly during the initial stage. This yields an even worse performance than PtO.
Both observations justify the necessity of calibrating the risk vector during the training process.

\subsection{Ablation on Efficiency (RQ4)}

In this section, we empirically evaluate the efficiency aspect of RTS-PnO, which is also deemed important in practical aspect~\cite{PTO-PNO-Benchmark}. Specifically, we evaluate both the training and inference efficiency. The experiment is conducted on five datasets: USD2CNY, USD2JPY, BTC, ETH, and S\&P 500. We report the result in mean and variable formats for training and inference per epoch.

\begin{figure*}[!htbp]
    \centering
    \subfigure[Total Training Time]{
    \begin{minipage}[t]{0.32\textwidth}
    \centering
    \includegraphics[width=\textwidth]{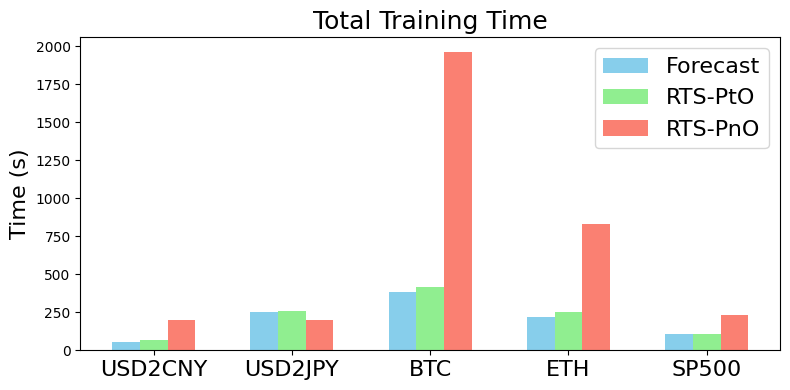}
    \label{fig:efficiency_total}
    \end{minipage}
    }
    \subfigure[Training Time per Epoch]{
    \begin{minipage}[t]{0.32\textwidth}
    \centering
    \includegraphics[width=\textwidth]{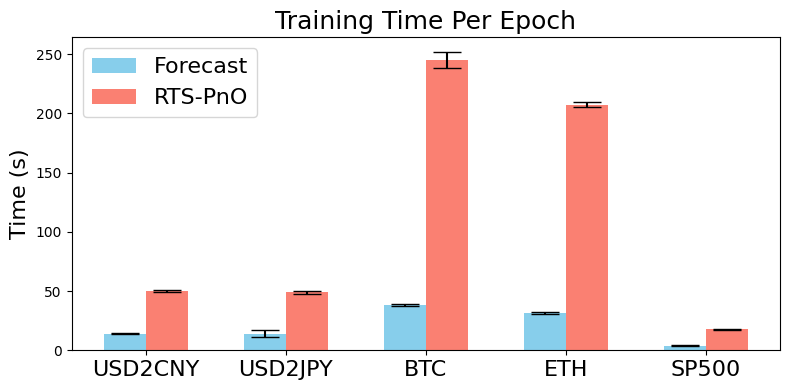}
    \label{fig:efficiency_train}
    \end{minipage}
    }
    \subfigure[Inference Time per Epoch]{
    \begin{minipage}[t]{0.32\textwidth}
    \centering
    \includegraphics[width=\textwidth]{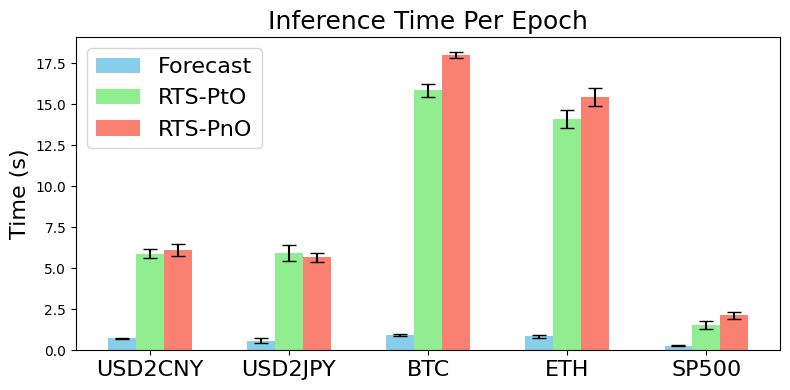}
    \label{fig:efficiency_inference}
    \end{minipage}
    }
    \caption{Efficiency Study on RTS-PtO and RTS-PnO.}
    \Description{Efficiency Study on RTS-PnO}
    \label{fig:efficiency}
\end{figure*}

For the training aspect, we report both total training time and training time per epoch, shown in Figures \ref{fig:efficiency_total} and~\ref{fig:efficiency_train}, respectively. We can observe that, compared with forecast methods, RTS-PnO does not require much additional training time. This suggests that the time required for the optimization process, compared with that for the training process, is rather neglectable. Additionally, when combining both Figures \ref{fig:efficiency_total} and~\ref{fig:efficiency_train}, we can observe that RTS-PnO requires much more training time per epoch, but the total training time is relatively at the same level compared to Forecast methods and RTS-PnO in certain cases. This suggests that RTS-Pno can converge to optimal with comparably fewer steps.

For the inference aspect, we can observe from Figure~\ref{fig:efficiency_inference} that RTS-PnO and RTS-PtO require much more time than Forecast methods. As these two frameworks require solving the optimization process during the inference time, it suggests that the optimization process dramatically slows down the inference speed. This is quite different from the previous observation that the time required for the optimization phase can be neglected compared to training. Such an observation calls for quicker optimization approaches, such as model-based optimization~\cite{MBO-Survey}.

\subsection{Investigation in RTS-PnO from Forecasting Performance (RQ5)}

In this section, we evaluate the forecasting performance of the forecasting model trained under the classic prediction paradigm and the proposed RTS-PnO framework. Note that the two-stage solution RTS-PtO has identical forecasting performance compared with the basic forecasting model, as it trains the same forecasting model during the prediction step, and the optimization step does not involve any update of the learned parameters. We report the performance on all three datasets. Both MSE and MAE are adopted to evaluate the forecasting performance, following the custom of time series forecasting~\cite{DLinear,PatchTST,FEDformer,TimesNet}. 

\begin{table}[!htbp]
    \centering
    \caption{Experiment on Forecasting Metrics}
    \vspace{-10pt}
    \label{tab:forecast}
    \begin{tabular}{c|c|cc|cc}
    \hline
        \multirow{2}{*}{Category} & \multirow{2}{*}{Dataset} & \multicolumn{2}{c|}{Prediction} & \multicolumn{2}{c}{RTS-PnO} \\
    \cline{3-6}
        & & MSE & MAE & MSE & MAE \\
    \hline
        \multirow{4}{*}{Currency}  
        & USD2CNY & \textbf{0.0049} & \textbf{0.0397} & \underline{0.0053} & \underline{0.0430} \\
        & USD2JPY & \textbf{0.0383} & \textbf{0.1263} & \underline{0.1201} & \underline{0.2796} \\
        & AUD2USD & \textbf{0.0277} & \textbf{0.1220} & \underline{0.0350} & \underline{0.1439} \\
        & NZD2USD & \textbf{0.0233} & \textbf{0.1072} & \underline{0.0327} & \underline{0.1334} \\
    \hline
        \multirow{2}{*}{Stock}
        & S\&P 500  & \textbf{0.1533} & \textbf{0.2744} & \underline{0.5567} & \underline{0.6194} \\
        & Dow Jones & \textbf{0.1184} & \textbf{0.2354} & \underline{0.3552} & \underline{0.4815} \\
    \hline
        \multirow{2}{*}{Criptos} 
        & BTC & \textbf{0.0197} & \textbf{0.0962} & \underline{0.0953} & \underline{0.2321} \\
        & ETH & \textbf{0.0213} & \textbf{0.1003} & \underline{0.1297} & \underline{0.2608} \\
    \hline
    \end{tabular}
\begin{tablenotes}
\footnotesize
\item[1] Here \textbf{bold} font indicates the best-performed method and \underline{underline} font indicates the second best-performed method. 
\end{tablenotes}
\end{table}
Based on the result in Table~\ref{tab:forecast}, we can easily witness a drop in both MAE and MSE after employing RTS-PnO. Such an observation aligns with previous works~\cite{DFL-Survey,PTO-PNO-Benchmark} that the PnO framework improves the decision quality at the cost of prediction accuracy. It also reveals the misalignment between the decision and prediction tasks. However, given the problem setup, the decision tasks play a more important role in the model training. 

\subsubsection{Potential Amid for reduced forecasting performance}
Our observation in the above section is not uncommon. It is commonly observed that the PnO framework could lower the prediction performance compared to purely prediction modules~\cite{PTO-PNO-Benchmark,DFL-Survey}. Previous researcher tends to focus on the decision quality. The prediction accuracy is also an important metric in evaluating the whole framework. One naive solution for solving the decreased prediction accuracy is to instead view prediction and optimization as a multi-task learning where the prediction and optimization module is the same as the PnO framework, and one additional calibration module is introduced to calibrate the prediction result. The prediction loss is applied to the calibrated result only, instead of the prediction result. General~\cite{MTL-Survey} and domain-specific~\cite{TS-OneFitsAll,UniTS} MTL frameworks can be borrowed as potential solutions.
\section{Online Experiment}
\label{sec:online}

\begin{figure}[!htbp]
    \centering  
    \includegraphics[width=0.95\columnwidth]{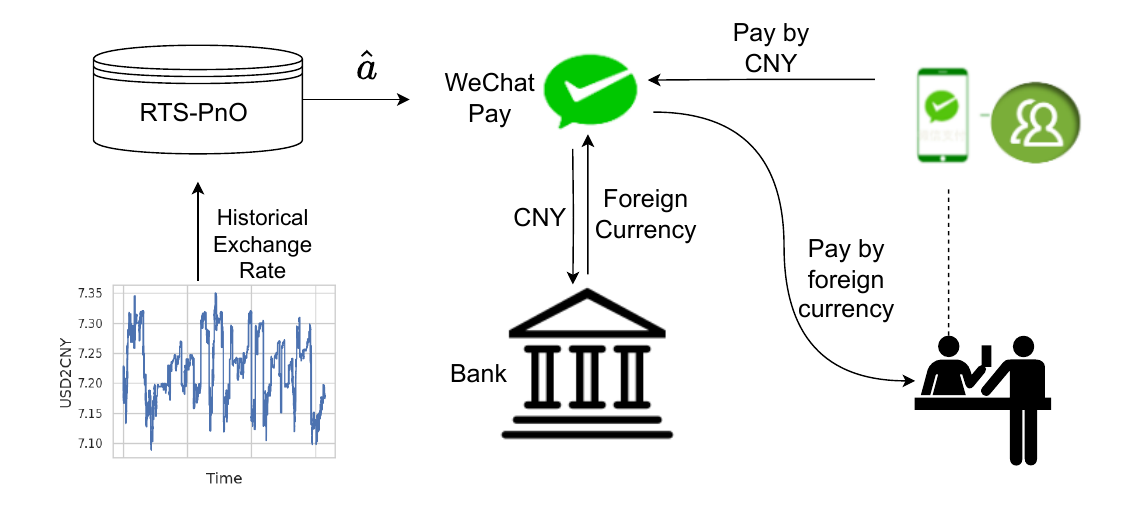}
    \caption{Application of RTS-PnO in the WeChat Pay's Cross-Border Payment Scenario.}
    \Description{Application of RTS-PnO in the Cross-Border Payment Scenario.}
    \label{online}
\end{figure}

We have deployed the proposed RTS-PnO at Tencent's financial platform to support WeChat Pay's cross-border payment scenarios. With the rapid increase in the number of Chinese residents traveling abroad, WeChat's cross-border payment service has now extended to millions of overseas merchants, facilitating the consumption of Chinese tourists abroad. 

As illustrated in Figure~\ref{online}, when consumers make purchases at overseas merchants, the amount in Chinese Yuan (CNY) that consumers are required to pay is calculated based on the exchange rate at the time of the transaction and is deducted from the consumer's account. Our platform then acquires the corresponding foreign currency amount for the user's expenditure and settles it to the merchant's bank account within a specified time window. To effectively manage funding costs, we employ our proposed RTS-PnO to allocate transaction funds within the designated time frame.  For example, consider a consumer who makes a purchase of \$100 USD at an overseas merchant when the exchange rate is 6.5 CNY/USD. In this scenario, the consumer will pay 650 CNY. Our platform will then acquire the \$100 USD within the next 24-hour window and settle the amount to the merchant's bank account. During this period, the exchange rate may fluctuate. To address this, we utilize the RTS-PnO framework to predict and determine the transaction volume to be exchanged each hour, thereby effectively managing funding costs. During the online experiment, no user data is involved. All involved data are obtained from third parties.

\begin{figure}[!htbp]
    \centering
    \includegraphics[width=0.95\columnwidth]{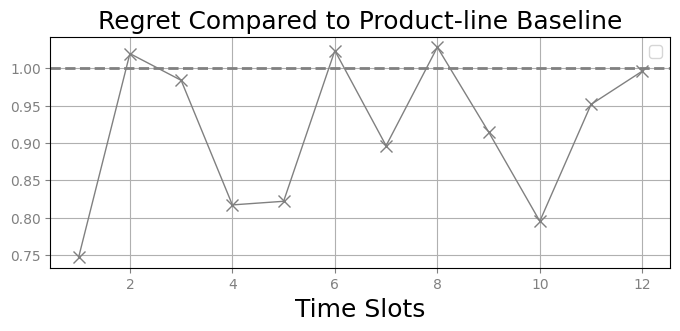}
    \vspace{-10pt}
    \caption{Online Result Compared with Product-line Baseline. The result is measured by the Regret compared with that of the Product-line Baseline. The dashed line represents the Product-line Baseline.}
    \Description{Online Result Measured by Relative Regret}
    \label{fig:online-result}
\end{figure}

We conducted an online experiment throughout 12 time slots in 2024, during which 50\% of the transactions were completed using the PTO framework, while the remaining 50\% utilized the RTS-PnO framework. The results show that the regret by RTS-PnO is only 91.6\% than that by product-line baseline, indicating that our proposed framework was able to reduce the average regret by 8.4\%. The detailed performance over each time slot is shown in Figure~\ref{fig:online-result}. It is important to note that this experiment is conducted when acquiring the WeChat Pay cash reserve. It does not affect the amounts paid by the consumers and received by the merchants in any circumstances.

\section{Conclusion}
\label{sec:conclusion}
In this paper, we study the fund allocation problem extended over the time dimension. Specifically, the target is to acquire a certain amount of asset within a period, and the price of the unit asset varies from time to time. We propose two solutions: (i) a two-stage solution RTS-PtO with fixed uncertainty constraint and (ii) an end-to-end solution RTS-PnO with adaptive uncertainty constraint. Both constraints are introduced to combine the hard-to-predict nature of time series and improve the decision quality. The evaluation is conducted over eight datasets from three categories of financial data. The proposed methods yield SOTA performance over the majority of the cases. Additionally, the online evaluation conducted over the cross-border payment scenarios of WeChat Pay demonstrates its feasibility in the real world.

\section*{Acknowledgement}

This research project is partially funded by the support of the SZTU University Research Project
(No.20251061020002). This work is done when Fuyuan, Xiuqiang and Xing are working as research intern and full-time employees at FiT, Tencent, respectively.

\section*{Ethical} In the online scenario involved in this paper, no user data is involved. The goal of the online task is to generate an allocation plan for buying currency at different time steps. During currency acquisition, all involved data are obtained from third parties (e.g., banks), and privacy regulations are satisfied. After acquiring the currency, the exchange rates for users are uniform and transparent. Neither process evolves any user's privacy data.

\section*{Limitations}
Though this paper empirically demonstrates the effectiveness of the proposed RTS-PnO in various benchmarks, this paper inevitably has certain limitations. 
First, the efficiency is a drawback of RTS-PnO. As illustrated in Figure~\ref{fig:efficiency_inference}, the inference time of RTS-PnO is larger than Forecasting-Only methods. Whether this inference increase is largely dependent on the scenarios. In our online setup, this is not a severe issue, as our online scenarios only require a 10-minute-level decision. However, in other scenarios, such as high-frequency trading, this inference increase might be an issue. Whether RTS-PnO satisfies the efficiency requirements is a domain-specific problem. The practitioners who aim to utilize RTS-PnO are advised to use their own judgment to conclude.
Second, the current RTS-PnO is limited to single-variable time-series benchmarks. The evaluation is not conducted on multi-variable time-series benchmarks. However, we believe it can be extended to multi-variable time-series benchmarks.

\normalem
\bibliographystyle{ACM-Reference-Format}
\balance
\bibliography{main}

\appendix

\section{Ablation Study on the Hyper-parameter Sensitivity}

\begin{figure}[!htbp]
    \centering
    \subfigure[USD2CNY]{
    \begin{minipage}[t]{0.22\textwidth}
    \centering
    \includegraphics[width=\textwidth]{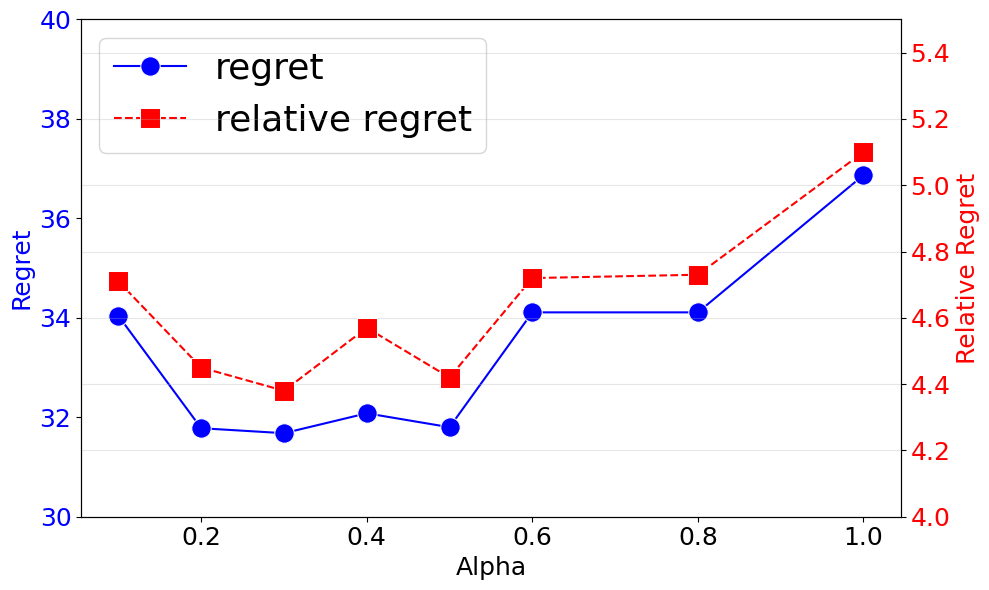}
    \label{fig:sensi-USD2CNY}
    \end{minipage}
    }
    \subfigure[D\&J]{
    \begin{minipage}[t]{0.22\textwidth}
    \centering
    \includegraphics[width=\textwidth]{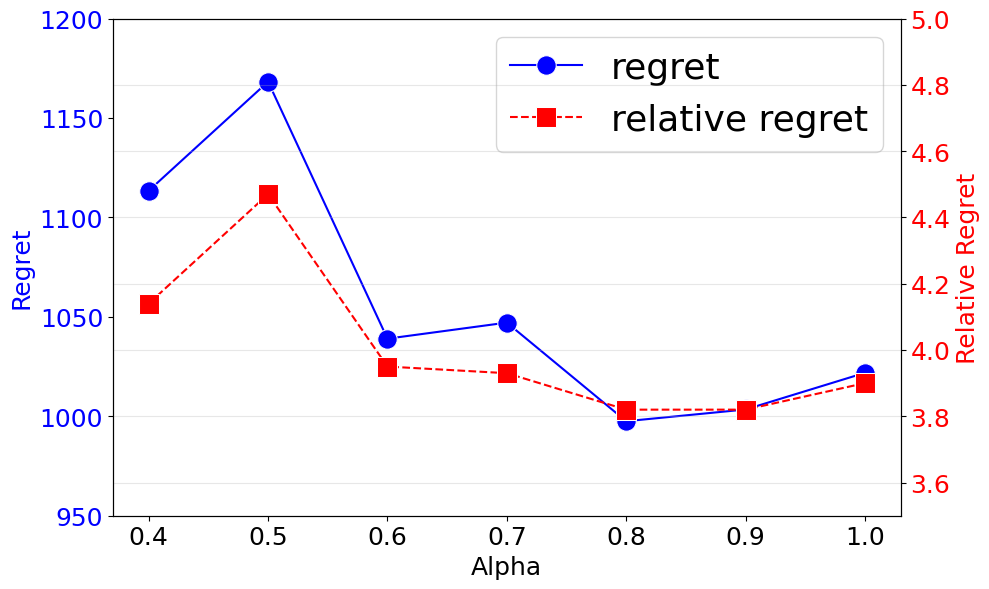}
    \label{fig:sensi-D&J}
    \end{minipage}
    }
    \subfigure[ETH]{
    \begin{minipage}[t]{0.22\textwidth}
    \centering
    \includegraphics[width=\textwidth]{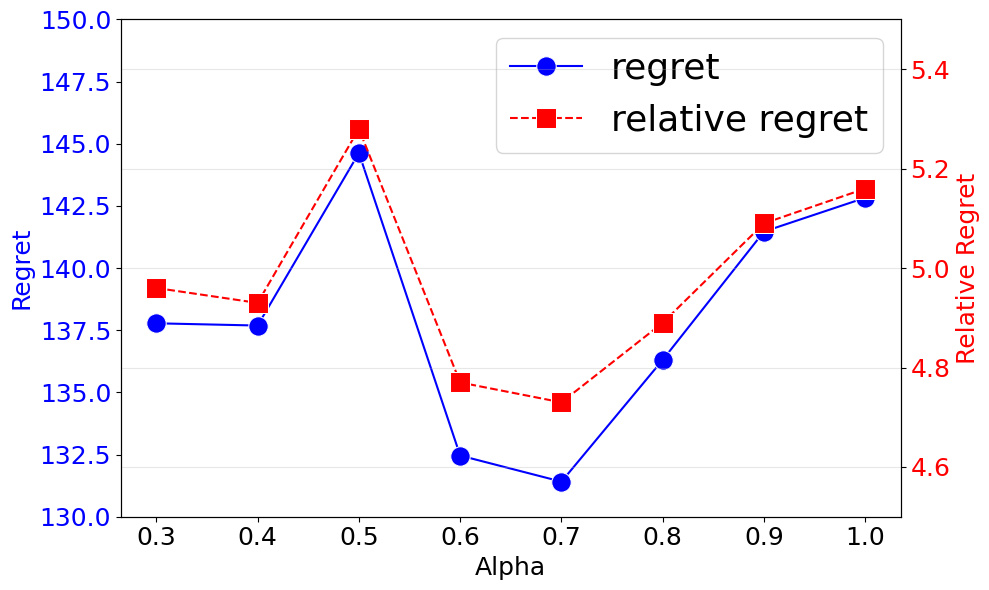}
    \label{fig:sensi-ETH}
    \end{minipage}
    }
    \caption{Sensitivity Study on $\alpha$.}
    \Description{Sensitivity Study on $\alpha$}
    \label{fig:sensi}
\end{figure}

\begin{figure}[!htbp]
    \centering
    \subfigure[USD2CNY]{
    \begin{minipage}[t]{0.45\textwidth}
    \centering
    \includegraphics[width=\textwidth]{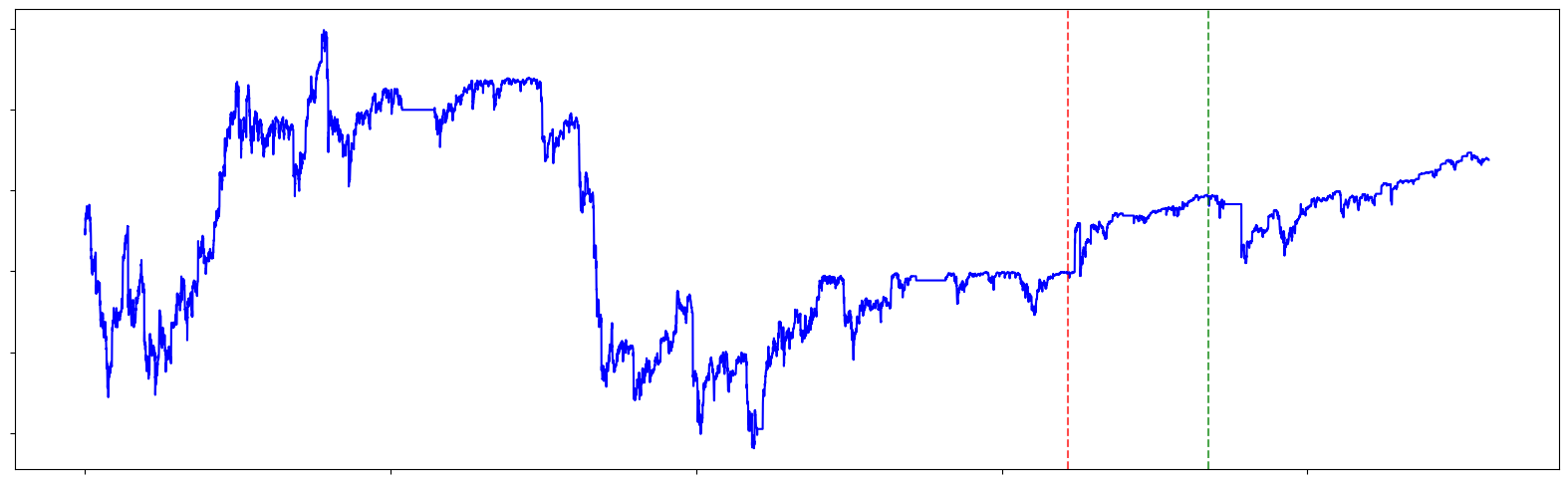}
    \label{fig:trend-USD2CNY}
    \end{minipage}
    }
    \subfigure[D\&J]{
    \begin{minipage}[t]{0.45\textwidth}
    \centering
    \includegraphics[width=\textwidth]{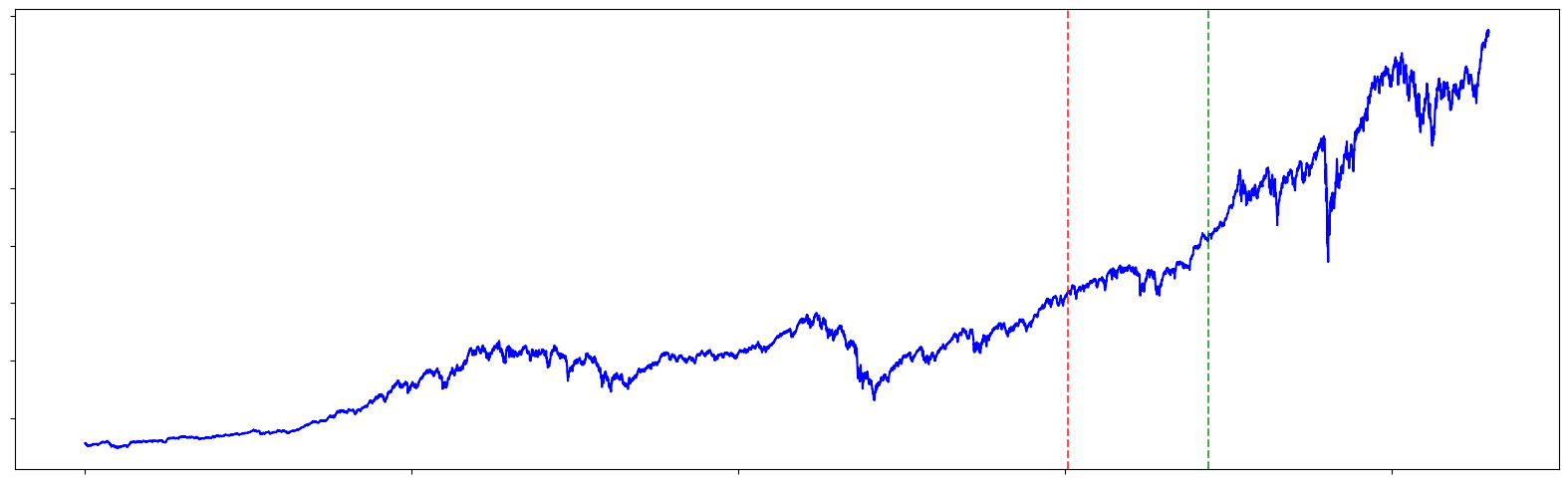}
    \label{fig:trend-D&J}
    \end{minipage}
    }
    \subfigure[ETH]{
    \begin{minipage}[t]{0.45\textwidth}
    \centering
    \includegraphics[width=\textwidth]{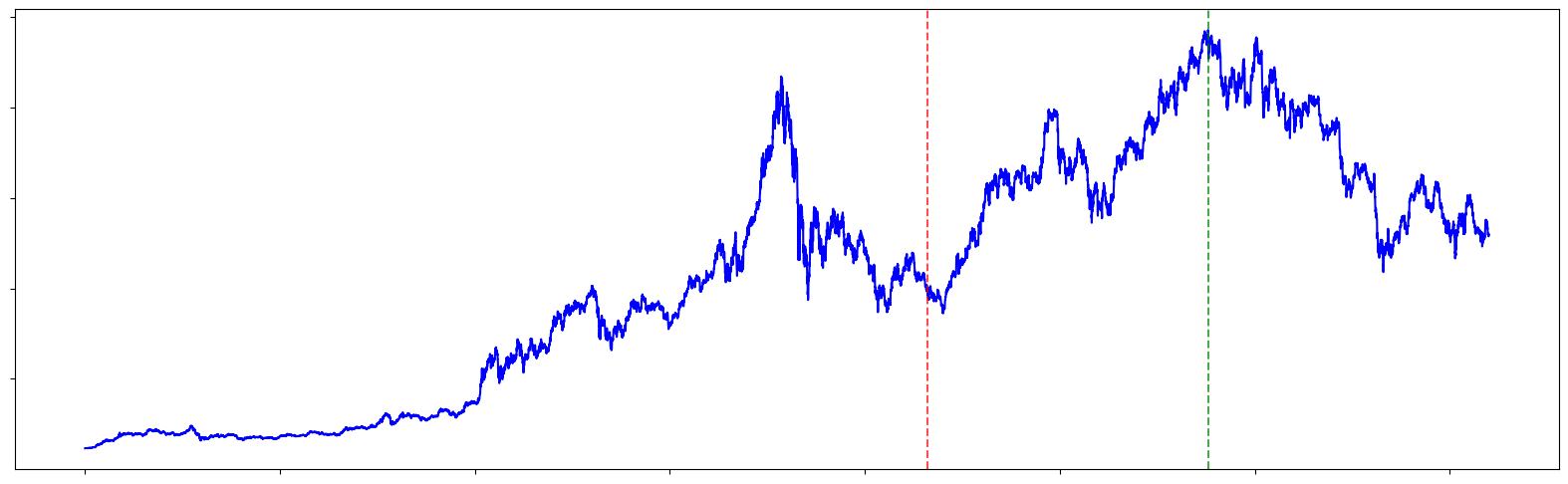}
    \label{fig:trend-ETH}
    \end{minipage}
    }
    \caption{General Trend of Related Datasets. Red and Green lines indicate the margin between training VS validation and validation VS testing sets.}
    \Description{General Trend of related datasets}
    \label{fig:trend}
\end{figure}

In this section, we study the sensitivity of $\alpha$ (also $r_0$) over three benchmarks: USD2CNY, D\&J and ETH in Figure~\ref{fig:sensi}. We also visualize the general trend of the involved datasets in Figure~\ref{fig:trend} to better illustrate how to select the optimal $\alpha$ on different benchmarks.

Given the observation in Figure~\ref{fig:sensi}, we can conclude that the optimal depends on the dataset. Generally speaking, a smaller $\alpha$ limits the position of feasible time slots. Currency data, such as USD2CNY, is changing relatively smoothly across time. This can be observed in Figure~\ref{fig:trend-USD2CNY}. It can be implied from Table~\ref{tab:statistics} that the max and min of Currency data are of relatively the same magnitude. So a smaller $\alpha$ can reduce the amount of unreliable predictions being considered during optimization. In contrast, the Stock and Cryptos datasets, represented by D\&J and ETH, may vary a lot across time and generally show a stronger trending behaviour, as shown in Figure~\ref{fig:trend-D&J} and~\ref{fig:trend-ETH} respectively. So a smaller $\alpha$ may eliminate all feasible choices during optimization in the extreme cases. Therefore, we need a larger $\alpha$ to loosen the constraint on prediction uncertainty. 

\end{document}